%% file: main.tex
\documentclass{ecai} 

\usepackage[ruled,linesnumbered]{algorithm2e}
\usepackage{multirow}
\usepackage{latexsym}
\usepackage{amssymb}
\usepackage{amsmath}
\usepackage{amsthm}
\usepackage{booktabs}
\usepackage{enumitem}
\usepackage{graphicx}
\usepackage{color}
\usepackage{array}
\usepackage{makecell}
\usepackage{bm}

\newcolumntype{P}[1]{>{\centering\arraybackslash}p{#1}}

\newcommand{\BibTeX}{B\kern-.05em{\sc i\kern-.025em b}\kern-.08em\TeX}

\begin{document}


\begin{frontmatter}




\title{HeLM: Highlighted Evidence augmented Language Model for Enhanced Table-to-Text Generation}


\author[A,D]{\fnms{Junyi}~\snm{Bian}\thanks{Email: 20110240003@fudan.edu.cn}}
\author[B]{\fnms{Xiaolei}~\snm{Qin}}
\author[B]{\fnms{Wuhe}~\snm{Zou}}
\author[B]{\fnms{Mengzuo}~\snm{Huang}}
\author[C]{\fnms{Congyi}~\snm{Luo}}
\author[E]{\fnms{Ke}~\snm{Zhang}}
\author[B]{\fnms{Congyi}~\snm{Luo}}
\author[B]{\fnms{Weidong}~\snm{Zhang}\thanks{Corresponding Author}}

\address[A]{School of Computer Science, Fudan University, Shanghai 200433, China}
\address[B]{Netease Games AI Lab, HangZhou, China}
\address[C]{School of Data Science, Fudan University, Shanghai 200433, China}
\address[D]{Shanghai Key Lab of Intelligent Information Processing, Fudan University, Shanghai 200433, China}
\address[E]{ShanghaiTech University, China}



\input{sections/s0_abstract}

\end{frontmatter}

\definecolor{gg}{RGB}{0,139,116}

\input{sections/s1_introduction}

\input{sections/s2_relatedwork}
\input{sections/s3_methodology}

\input{sections/s4_experiments}

\input{sections/s5_conclusion}

\bibliography{main}

\end{document}

%% file: sections/s0_abstract.tex



\begin{abstract} 

Typically, Table-to-Text generation models directly expand tables line by line into a long string as input makes it difficult for generators (Language Models) to focus on saliency units or data units relevant to the query from the unstructured, lengthy strings. To address this issue, we propose a two-step solution called HeLM which consists of two modules: a table highlighter that identifies relevant row evidence, and a table summarizer that generates sentences based on the highlighted table. To facilitate this, we utilized the open-sourced large language model LLaMA2 as the backbone for these two modules and efficient finetuning on it. We also propose a searching algorithm along with label distillation to construct highlighting labels for obtaining the table highlighter. On both the FetaQA and QTSumm datasets, our approach achieved state-of-the-art results in ROUGE and BLEU scores. Additionally, experiment results show that highlighting the key evidence on input tables significantly enhances the model's performance and provides valuable interpretability.

\end{abstract}



%

%% file: sections/s1_introduction.tex
\section{Introduction}
Tabular data represents a common form of structured information within the realm of computing, particularly within databases. Analyzing and manipulating tabular data holds significance in the domain of Natural Language Processing (NLP).
A practical scenario for automated table processing involves posing a query or requirement to the table, which can take the form of questioning \cite{2015_wtq,2022_fetaqa} or summarization \cite{2023_qtsumm}, among others. Subsequently, based on this demand, a response is generated, typically in the form of sentences or paragraphs. This type of task is referred to as table-to-text generation.
The recent emergence of Large Language Models (LLMs) \cite{2020_llms1,2023_gpt4,2022_cotllm2,2023_llama,2022_scaling,2022_bloom} showcase impressive capabilities in multiple NLP tasks, unveiling vast potential in handling tabular data. Therefore, this paper delves specifically into the application of LLMs in table-to-text generation tasks, exploring the immense potential of LLMs in tabular data generation through relevant research and empirical analysis.

Recent approach \cite{2023_tabcot,2023_dater} in utilizing LLMs for table-based tasks usually relies on invoking online APIs
for few-shot learning or integrated techniques such as chain-of-thought \cite{2022_cotllm2} or in-context learning, achieving comparable performance even without fine-tuning. However, these methods necessitate frequent API calls and, in practical application, entail uploading table data, thereby posing a risk of information leakage.
Therefore, a finetuned LLM system specialized in handling tabular data stands as an effective solution.
With the availability of open-sourced LLMs\cite{2022_glm,2023_llama,2023_llama2,2023_Falcon,2023_mistral} and the introduction of parameter-efficient training methods\cite{2021_lora,2023_qlora}, fine-tuning a large language model with limited computational resources is now available. Therefore, in this study, we employ QLoRA \cite{2023_qlora} to fine-tune the LLaMA2 \cite{2023_llama2} base model specifically for Table-to-Text generation.

To enable the model to adeptly handle tabular data, it requires the capability to reason intricately across textual, numerical, and logical domains. 
Some methods \cite{2017_tabrea_sql_01, 2022_tabres_sql_02} achieve this by synthesizing executable languages, such as SQL. Others \cite{2020_tapas, 2021_tapex, 2022_omnitab, 2022_pasta} pretrain on additional table data to acquire table reasoning capabilities. However, most LLMs often lack table reasoning capabilities due to their pretraining text containing minimal tabular data content.
In this paper, we conceptualize table reasoning as the capacity to identify crucial evidence within a table according to the output requirements. In this context, we define evidence as the specific row-level data crucial for answering the final output. Considering that input tables are often extensive, essential information usually resides within a small portion. Identifying and conveying these row data effectively to the model can significantly enhance the model's output quality.

In real-world scenarios, input of table-to-text generation often consists solely of entire tables and queries, necessitating an automated process to gather evidence data. 
This paper introduces a two-step methodology designed to tackle these challenges. 
The first is an LLM-based table highlighter, aimed at identifying and \textbf{H}ightlighting \textbf{e}vidence given input table. Then another Large \textbf{L}anguage \textbf{M}odel based table summarizer model acquiring highlighted table as input and generates the final output. This methodology is termed as \textbf{HeLM}.



The pivotal component of HeLM lies in the table highlighter, which outputs evidence(relevant row indexes) based on the given table and query. However, most datasets lack evidence labels, making the fine-tuning of LLMs inconvenient.
To address this issue, we propose two methods for obtaining evidence labels.
One direct approach is to distill evidence labels from more powerful LLMs. Additionally, we also designed an algorithm that, without relying on distillation, automatically constructs evidence labels using only the original input-output data from the dataset. After that, combining evidence labels obtained from different methods can further enhance the quality of evidence. The table highlighter trained in this manner not only improves the overall performance of HeLM but also provides valuable interpretability.
The contributions of this paper are as follows: 

\begin{itemize}
    \item We propose a two-step table-to-text approach named HeLM, which utilizes a table highlighter to highlight input tables, aiding downstream table summarizers in producing better text outputs.
    \item We introduce a search-based evidence label construction method and a workflow for training HeLM's highlighter and summarizer.
    \item HeLM attains state-of-the-art results in terms of BLEU and ROUGE scores on both the FetaQA and QTSumm datasets and the code\footnote{https://github.com/Eulring/HeLM} is released.
\end{itemize}

%% file: sections/s2_relatedwork.tex
\input{miscs/figure_framework}

\section{Related work}
\label{sec:relate}




\subsection{Reasoning Over Tables}

Enhancing a model's table reasoning capabilities is pivotal for table-related tasks. A prevailing strategy is to pre-train models on reasoning data that combines tables and text. Such pre-training methods can be encoder-based, as exemplified by Tapas \cite{2020_tapas}, which pre-trains on a large-scale corpus of text-table pairs. TaBERT \cite{2020_tabert}, on the other hand, employs Masked Column Prediction and Cell Value Recovery as pre-training tasks to fine-tune a BERT-based table encoder. \citep{2022_turl} introduce the Masked Entity Recovery objective for pre-training to distill the semantics and knowledge within vast amounts of unlabeled data. There are also generative pre-training models like Tapex \cite{2021_tapex}, which utilize a novel SQL execution task to conduct table pre-training on a diverse, large-scale, and high-quality synthetic dataset. UnifiedSKG \cite{2022_unifiedskg} transforms various types of structured data tasks into a unified text-to-text format for joint training, including those involving tabular data. 
TableGPT\cite{2023_tablegpt} fine-tuned on LLM while utilizing an accompanying table encoder to attain a comprehensive understanding of the input table.

The above models often gain table reasoning capabilities in an end-to-end manner, sacrificing explainability. 
Some studies first use explicit methods for table reasoning, before proceeding with generation.
The approach, named by REFACTOR \cite{2023_qtsumm}, suggests generating query-relevant facts from tables as intermediate results for LLM's input. Another noteworthy method\cite{2022_binding}, employs Codex \cite{2021_codex} to synthesize SQL for executing logical forms against tables in question-answering tasks. 
Dater \cite{2023_dater} takes an approach by reducing the original table into relevant sub-tables and ask LLMs to execute query-focused SQL language to get numerical context.
Our approach employs a highlighter first for explicit table reasoning, followed by generation. The reasoning capability of the highlighter stems from tabular data and distillation.
 


\subsection{Instruction Tuning}

Instruction tuning \cite{2022_it1, 2024_it2, 2021_it3} is a learning paradigm that converts conventional supervised learning datasets into an instruction-following format, By learning the corresponding outputs for instructions, thereby guiding the output of LLMs. Recent breakthroughs \cite{2023_it_dis1, 2023_it_dis2, 2023_it_dis3} have also facilitated the development of smaller models demonstrating task-adherence capabilities, achieved through fine-tuning on instruction data generated by LLMs, such as ChatGPT or GPT4 \cite{2023_gpt4}.

Instruction fine-tuning is typically applied in general NLP tasks \cite{2022_it4}. Despite some methods \cite{2023_dater, 2023_tabcot} exploring instruction design for direct LLM inference, the utilization of instruction tuning in table-to-text generation remains underexplored.


%% file: miscs/figure_framework.tex
\begin{figure*}[htp]
    \centering
    \includegraphics[width=1.0\textwidth]{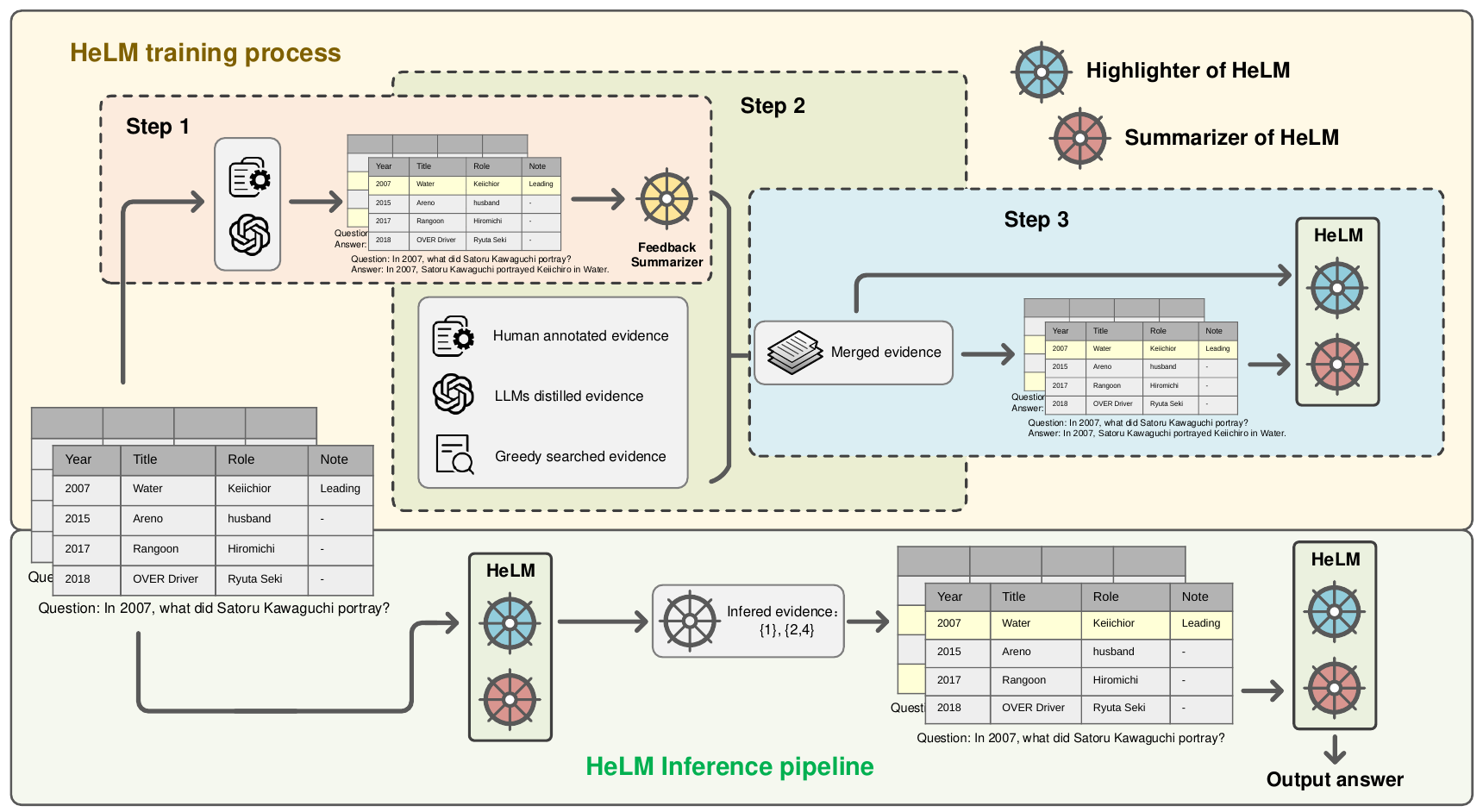}
    \caption{The overall framework of HeLM. The upper part demonstrates the training process, while the lower part illustrates the inference process.}
    \vspace*{0.4cm}
    \label{fig:framework}
\end{figure*}

%% file: sections/s3_methodology.tex
\section{Methodology}

\subsection{Task Formulation}

In table-to-text generation datasets, the input consists of a structured Table \( T \) containing rows and columns of data, along with a natural language question \( Q \). The golden output is denoted as \( \mathcal{Y} \).
A table-to-text generation system is required to generate a response \(\hat{\mathcal{Y}}\) that answers the question, leveraging the information encoded within the table.


\input{miscs/figure_prompt_hellama}
\input{miscs/figure_prompt_distill}

\subsection{HeLM Framework}
Our framework consists of two components: a table highlighter $\mathcal{M}_{H}$ and a table summarizer $\mathcal{M}_{S}$. 
As shown in the Figure\ref{fig:framework}, the table highlighter identifies the indexes of row data relevant to the query within the table. 
For this component, we employ an LLM that directly generates index numbers as text.
To achieve this, we design a $\text{Prompt}_H$ that concatenates rows of the table into a string along with the query and task description, forming the input for the table highlighter $\mathcal{M}_{H}$. the details of the $\text{Prompt}_H$ is in Figure\ref{fig:prompt_helm}.
In training, golden evidence is incorporated into the \( \text{Prompt}_H \). At inference, the \textit{golden output} is set to empty:

\begin{align}
    E = \mathcal{M}_{H} \left( \text{Prompt}_H(T, Q) \right) 
\end{align}

The output \(E = \{e_i, ...\}\) of \(\mathcal{M}_{H}\) is a list of indices, where \(e_i\) corresponds to the row number in the input table $T$.

\input{miscs/figure_hl}
After obtaining the evidence indexes, we become aware of which row data in the original table $T$ is relevant to the query $Q$. We highlight this evidence on $T$ to produce the modified table $T^{*}$. The highlighting operation $\text{HL}(\cdot)$ involves adorning each data cell of a key row with a distinctive `*' character, signifying its importance, an exmple is illustrated in Figure \ref{fig:hl}.
\begin{align}
\label{eqn:highlight}
    T^{*} = \text{HL}(T,E)
\end{align}

The table summarizer will subsequently produce the final result based on the prompt generated by highlighted table $T^{*}$ associated with query $Q$ and task description:
\begin{align}
    \hat{\mathcal{Y}} = \mathcal{M}_{S} \left( \text{Prompt}_S(T^{*}, Q) 
 \right)
\end{align}

\subsection{Evidence Labels}
Training a good highlighter necessitates high-quality evidence labels. Prior to delving into evidence label searching and evidence label merging, we first introduce Evidence Feedback.

\subsubsection{Evidence Feedback}
To distinguish the quality of an evidence label, we require a feedback model denoted as $\mathcal{M}_F$. 
Given an evidence $E$. Upon combining the evidence with the table and using it as input, the feedback model $\mathcal{M}_F$ will produce an output $\hat{\mathcal{Y}}$:

\begin{align}
    \hat{\mathcal{Y}} = \mathcal{M}_F (\text{Prompt}_S(HL(T,E), Q) )
\end{align}



The quality of the evidence is assessed based on the goodness of this output:
\begin{align}
    Reward = eval(\mathcal{Y}, \hat{\mathcal{Y}})
\end{align}

Alternatively to highlighting within the table, there is a more sensitive feedback approach that directly extracts row data from the table based on the evidence, resulting in a sub-table, which we define as $SubTab(T, E)$.


Essentially, the function of the feedbacker \( \mathcal{M}_F \) and the summarizer \( \mathcal{M}_S \) is the same; both return results based on a table that has been modified by evidence. However, the performance requirements for \( \mathcal{M}_F \) are not high; it only needs to be sensitive to the evidence, meaning that results derived from good evidence should be significantly better than those from poor evidence. Before obtaining a high-quality \( \mathcal{M}_S \), we can utilize limited resources to first train an \( \mathcal{M}_F \).

\subsection{Evidence Labels Construction}

\label{sec:elabel}
Query-focused evidence indexes are necessary for fine-tuning a table highlighting module, and we summarize three sources for obtaining these evidence labels.
\begin{itemize}
    \item \textbf{Human annotated evidence} $E_{manul}$: Some datasets, such as QTSumm \cite{2023_qtsumm}, inherently include labels for relevant evidence, and they are obtained through manual annotation.
    \item \textbf{Distilled evidence} $E_{distill}$: Labels can also be distilled from other LLMs such as GPTs \cite{2023_gpt4}. To better capture evidence, we designed an few-shots in-context learning prompt(see Figure\ref{fig:prompt_distill}), incorporating golden labels $\mathcal{Y}$ to better capture evidence.
    \begin{align}
        E_{distill} = \text{LLMs}(\text{Prompt}_{distill}(T, Q, \mathcal{Y}, \text{Examples}))
    \end{align}
    \item \textbf{Searched evidence} $E_{search}$: Evidence labels can also be obtained through search algorithms, which require feedback for different $E$. This feedback system has two requirements: one is the golden output $\mathcal{Y}$ corresponding to the input table and query, and the other is a feedback table summarizer $\mathcal{M}_{F}$. For more details of this algorithm, please refer to section \ref{sec:labelsearch}.
    \begin{align}
        E_{search} = \text{Search}(T, Q, \mathcal{Y}, \mathcal{M}_{F})
    \end{align} 
\end{itemize}


The table evidence labels obtained through various methods showcase significant disparities. By integrating these evidence labels, higher-quality evidence can be attained. This process entails using highlighted tables associated with different evidence and getting sentences via the feedbacker $\mathcal{M}_F$. The evidence label for the current sample is chosen based on the sentence that receives the highest evaluated score. The formula for generating the merged label $E_{merge}$ is outlined as follows:

\begin{align}
\label{eqn:merge}
    E_{merge} = \text{Merge}(\bm{E}, T, Q, \mathcal{Y}, \mathcal{M}_{F})
\end{align}
Here, $\bm{E}$ represents the available evidence label set. For datasets lacking human annotated evidence, $\bm{E}=\{E_{search}, E_{distill}\}$.

\subsection{Evidence Labels by Searching}
\label{sec:labelsearch}
As mentioned earlier, the search algorithm requires a feedback system. The feedback system includes the golden output $\mathcal{Y}$ corresponding to the table query and a feedbacker $\mathcal{M}_F$. 
$\mathcal{M}_F$'s output is evaluated by computing the BLEU score against $\mathcal{Y}$ to derive numerical feedback value.


In label searching, the input for $\mathcal{M}_F$ is the sub-table corresponding to the evidence.
The reason for using the sub-table as input to search for evidence is that $\mathcal{M}_F$ is more sensitive to sub-table evidence compared to the input of the complete table with evidence highlighted. 
Because even when relevant row data is not highlighted as evidence in the complete table input, $\mathcal{M}_F$ might still capture it.

Assuming the table has $n$ rows of data, and each row can be either selected or not, the search space for this algorithm is $2^n$. This implies that for each training example, one would need to invoke LLMs (summarizer) $2^n$ times to construct the optimal evidence, which is impractical. Therefore, we propose a greedy search method to construct labels, reducing the searching complexity from $O(2^n)$ to $O(n)$.

\input{miscs/algotirhm_gs}

The core idea of this algorithm is that query-relevant evidence can enhance the summarizer feedback score, while irrelevant evidence cannot. During evidence construction, the initial evidence is an empty set. Based on feedback results, we expand this evidence by adding row index one by one, and we repeat this process until the score no longer increases or reach a certain step. We have also designed heuristic steps to efficiently select evidence rows. A more detailed procedure is shown in Algorithm \ref{alg:gs}.

\subsection{HeLM Training}

HeLM comprises two modules: Highlighter and Summarizer. The training of the Highlighter utilizes the highest-quality evidence label $E_{merge}$. Training a complete HeLM modules involves the following steps:

\paragraph{Step 1. Obtain feedback summarizer:} Distilling $E_{distill}$ through LLMs, and training a rough table feedbacker/summarizer $\mathcal{M}_F$ using $E_{distill}$. 
\begin{align}
    \left\{ \text{HL}(E_{distill}, T), Q, \mathcal{Y} \right\} \rightarrow \mathcal{M}_F
\end{align}

\paragraph{Step 2. Obtain merged evidence:} Obtaining $E_{search}$ using Algorithm \ref{alg:gs}, then combining the existing evidence through Equation \ref{eqn:merge} to obtain $E_{merge}$.
 
\paragraph{Step 3. Fine-tuning highlighter and summarizer:} Train highlighter $\mathcal{M}_{S}$ using $E_{merge}$, and train summarizer $\mathcal{M}_{R}$ using $\mathcal{Y}$ and $T^*$ corresponding to $E_{merge}$.
\begin{align}
     \left\{ T^{*}, Q, \mathcal{Y} \right\} 
 \rightarrow \mathcal{M}_{S} \\
     \left\{T,Q,E_{merge} \right\}  \rightarrow \mathcal{M}_{H}
\end{align}
Figure \ref{fig:framework} displays the comprehensive training and inference process of HeLM.

 
Facing the immense size of recent LLMs, conducting full-parameters fine-tuning is prohibitively expensive. As a practical alternative, we adopt the parameter-efficient finetuning strategy, QLoRA \cite{2023_qlora, 2021_lora}, to train our highlighter and summarizer. 
This approach significantly reduces trainable parameters to 0.6\% of the original, enabling fine-tuning of LLMs on consumer devices.

%% file: miscs/figure_prompt_hellama.tex
\begin{figure}[h]
    \centering
    \includegraphics[width=\linewidth]{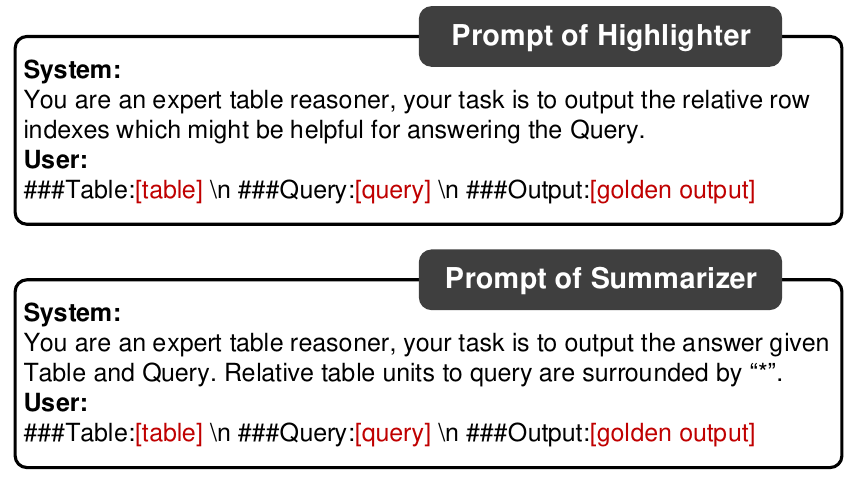}
    \caption{Prompt of Highlighter and Summarizer. The elements within the red brackets can be replaced based on different examples.}
    \vspace*{0.4cm}
\label{fig:prompt_helm}
\end{figure}

%% file: miscs/figure_prompt_distill.tex
\begin{figure}[h]
    \centering
    \includegraphics[width=\linewidth]{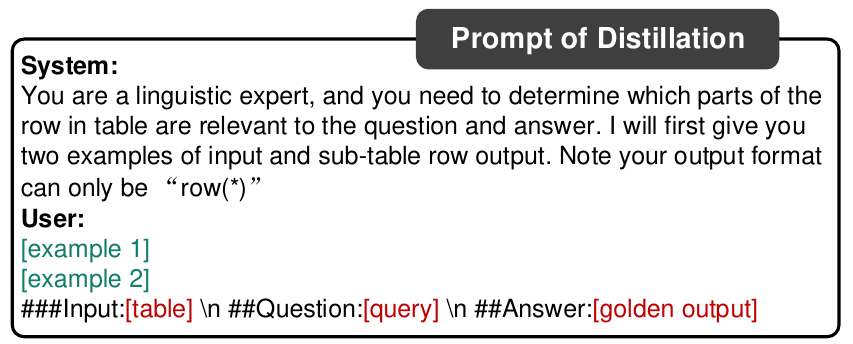}
    \caption{Prompt of evidence labels distillation.}
    \vspace*{0.4cm}
\label{fig:prompt_distill}
\end{figure}

%% file: miscs/figure_hl.tex
\begin{figure}[h]
    \centering
    \includegraphics[width=0.48\textwidth]{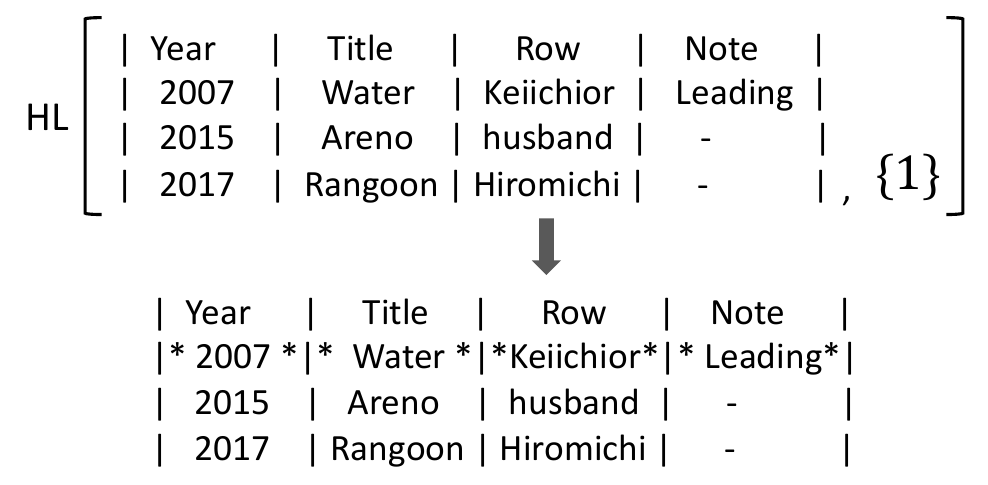}
    \caption{Case of table highlighting, \(\{1\}\) corresponds to \(E\) in equation \ref{eqn:highlight}, and the visualized table corresponds to \(T\). The output below is \(T^{\star}\).}
    \vspace*{0.4cm}
    \label{fig:hl}
\end{figure}


%% file: miscs/algotirhm_gs.tex
\begin{algorithm}[tb] \small
    \LinesNumberedHidden
    \caption{Reasoning evidence labels by search}
    \label{alg:gs}
    \KwIn{Table $T$($n$ rows), Query $Q$, Answer $\mathcal{Y}$, Feedback summarizer $\mathcal{M}_F$}
    \KwOut{Searched evidence Label $E_{search}$}
    Generate $n$ evidence labels $\bm{E} = \{ E_1, E_2,..., E_n \}$, where $E_i = \{i\}$  \\

    \For{$i \leftarrow 1$ to $n$}{
        $\hat{\mathcal{Y}}_i = \mathcal{M}_F(\text{Prompt}_S(\text{SubTab}(T, E_i), Q))$ \\
        $R_i = \text{eval}(\hat{\mathcal{Y}}_i, \mathcal{Y})$ \\
    }

    Reorder the $E$ according to reward $R$. \\

    Evidence label $E_s$ is initialized with empty set.\\

    Evidence label reward: $R_s = 0$ \\

    \For{$i \leftarrow 1$ to $n$}{
        $\mathcal{Y}_{i} = \mathcal{M}_F(\text{Prompt}_S(\text{SubTab}(T, E_i+E_s), Q))$ \\
        $R_{i} = \text{eval}(\mathcal{Y}_{i}, \mathcal{Y})$ \\
        \If{$R_{i} > R_s$}{
            $R_s = R_{i}$ \\
            $E_s = E_i + E_s$ \\
        }
    }
\end{algorithm}
    %

%% file: sections/s4_experiments.tex
\section{Experiments}
\label{sec:exp}

\subsection{Dataset and Evaluation}
\textbf{FeTaQA:}  FeTaQA is a dataset designed for free-form table question-answering, constructed using information from Wikipedia. It introduces a table question answering scenario, where questions are answered in natural language. The FeTaQA dataset comprises 7,326 question-answer pairs in the training set, 1,000 in the validation set, and 2,006 in the test set. For the evaluation of results on the FeTaQA dataset, we employ commonly adopted metrics, including ROUGE-1, ROUGE-2, ROUGE-L \cite{2004_rouge}, and the BLEU \cite{2002_bleu,2018_bleu} score.

\noindent \textbf{QTSumm:} QTSumm is a query-focused table summarization dataset, requiring text generation models to engage in human-like reasoning and analysis over the provided table to generate a tailored summary. The training and validation sets consist of 4,981 and 1,052 examples respectively, and the test set comprises 1,078 examples. Notably, in comparison to the FeTaQA dataset, QTSumm exhibits longer output lengths. For the evaluation of results on QTSumm, we employ not only ROUGE-L and BLEU scores but also the METEOR \cite{2005_meteor}  as the evaluation metric.

\input{miscs/table_fetaqa.tex}
\input{miscs/table_qtsumm.tex}

\subsection{Implementation Details}

 
All models are executed on a single NVIDIA-A100 GPU with 80G of memory. We optimized our baseline LLMs through 4-bit QLoRA finetuning, utilizing an effective batch size of 8 for 2 epochs. The optimization process employed the AdamW \cite{2018_adamw} optimizer with default momentum parameters and a constant learning rate schedule set at 2e-4. For QLoRA, NormalFloat4 with double quantization was applied to the base models, and LoRA adapters were added to all linear layers with parameters $r=16$ and $\alpha=32$. The maximum input length was constrained to 2048. For generating outputs from the LLMs, we employed nucleus sampling \cite{2019_sample} with parameters $p=0.9$ and a temperature of $0.1$.

Our model, $\text{HeLM-13B}$, denotes that both the summarizer and highlighter utilize Llama2-13b-hf\cite{2023_llama2} as the backbone model for parameter-efficient fine-tuning. 

\subsection{Baselines}


There are primarily two types of baselines for Table-to-Text task, that is, fine-tuning methods, and few-shot methods using LLMs. In the FeTaQA dataset, fine-tuning baselines contain the T5-based \cite{2020_T5} models (T5-Small, T5-Base, and T5-Large), as well as TAPEX \cite{2021_tapex}, OmniTab  \cite{2022_omnitab}, and PLOG \cite{2022_plog}. TAPEX and OmniTab are both BART-based models, with additional pre-training on custom training data. 

HeLM-13B refers to a model where both the highlighter and summarizer are fine-tuned on the basis of LLaMA2-13B in a parameter-efficient manner with the parameters used for training constituting merely 0.6\% of the LLaMA2. In contrast, we have also present the approach of directly parameter-efficient fine-tuning LLaMA2-13B. 



In FeTaQA dataset, methods using LLMs for few-shot learning include Dater (Codex) \cite{2023_dater} and TabCOT \cite{2023_tabcot}. The few-shot LLMs baselines for the QTSumm dataset are directly adapted from \cite{2023_qtsumm}, including methods such as LLaMA2-7B,13B,70B and GPT-3.5,4.

\subsection{Main Results}   
$\text{HeLM-13B}$ demonstrate superior performance on both the QTSumm and FeTaQA datasets. Specifically, on the FeTaQA dataset (see Table \ref{tab:fetaqa}), 
$\text{HeLM-13B}$ outperforms the previous leading method, Dater, with a $1.8$ and $1.9$ improvement in Rouge-1 and Rouge-L respectively. More notably, there is a substantial improvement in the BLEU score, with an increase of $3.26$. 



The results for the QTSumm dataset are presented in Table \ref{tab:qtsummm}. HeLM-13B achieved the best results in both the BLEU and ROUGE-L metrics, with improvements of $1.7$ and $2.5$ respectively over the second-ranked LLaMA2-13B. In terms of the METER metric, HeLM-13B ranked second, showing an improvement of $3.3$ over LLaMA2-13B, and was $1.1$ behind GPT-4. This demonstrates the effectiveness of highlighting mechanism.


\subsection{Human Evaluation}

Relying solely on the ROUGE and BLEU scores cannot comprehensively assess the model's performance. Therefore, human evaluation is necessary. In human evaluation, we recruited and paid three well-educated graduate students. Before starting the annotation, we wrote a unified annotation manual to guide the annotators on how to perform the annotations. The three annotators were required to rank the outputs of three methods on 100 samples. For each sample, we provided the input table, query, and the golden answer for reference.

We conducted a human evaluation in three aspects: (1) \textbf{Fluency} (whether the output sentences are fluent and without grammar errors). (2) \textbf{Correctness} (the accuracy of numerical values and logical correctness of sentences). (3) \textbf{Adequacy} (whether the output results cover all aspects of the questions). Compared models included LLaMA2-13B LoRA, which is also based on efficient fine-tuning, as well as Tabcot, an LLMs-based few-shot method. 
%
Among these three metrics, correctness is the most indicative table reasoning ability. In table \ref{tab:humaneval}, TabCot performs lower on fluency compared to LLaMA2-13B LoRA, but its correctness is significantly better. 
This suggests that fine-tuning on a specific dataset is more focused on learning surface-level features. 
Regarding the table reasoning ability as indicated by correctness, LLMs like GPT-3 showcase superior capabilities. HeLM performs best in correctness, indicating the positive impact of HeLM's highlighter on the overall accuracy of the results.

\input{miscs/table_humanEval}
\subsection{Ablation Study}

\subsubsection{Impact of Model Size}
As shown in Table \ref{tab:aba1}, when using LLaMA2-7B as the base model for fine-tuning, $\text{HeLM-7B}$ showed a $2.15$ decrease in BLEU score and a $2.5$ decrease in ROUGE-L score compared to HeLM-13B. Directly fine-tuning LLaMA2-7B using LoRA also exhibited a $1.4$ decrease in ROUGE-L compared to LLaMA2-13B. 


The phenomenon of model performance increasing with size is quite common. At the same time, we have also found that the performance improvement of HeLM with increased model size is significantly higher than that of LLaMA2. This underscores the importance of  table highlighting in HeLM. It is also suggested thst simply increasing the model size without considering the model's reasoning capabilities does not yield substantial returns.

\subsubsection{Impact of Table Highlighting}
HeLM's summarizer takes tables highlighted with evidence as input, and different evidence will have different effects on the output results of the summarizer. We record this experiment in Table\ref{tab:aba2}. 
When the summarizer of $\text{HeLM-13B}$ receives unmodified tables as input, specifically, 
the result of \textbf{-w/o highlight} showed a decrease of both BLEU and ROUGE-L. This signifies the effectiveness of highlighting crucial information in LLM's input tables. Additionally, when using the same evidence for input table, and constructing a sub-table with only key row information as input instead of retaining all table data, the approach \textbf{-subTab} has a $2.82$ decrease in BLEU score. This suggests the benefit of retaining sufficient table information. 
Another observation is that when no highlighting is applied to the input table, LLaMA2 outperformed HeLM-w/o HL. This happens because HeLM's summarizer generated dependency on highlighted evidence during training. However, during testing, when the highlighting is absent, it results in poorer performance compared to LLaMA2.

\input{miscs/table_fetaqa_aba1.tex}
\input{miscs/table_fetaqa_aba2.tex}

\subsubsection{Impact of Evidence Labels}
During evaluation, the evidence used for highlighting the input table in our base model HeLM is generated by the HeLM's highlighter. By keeping the summarizer of HeLM-13B fixed, we also examine the output derived from employing various evidence labels for table highlighting, aiming to illustrate the impact of evidence labels quality. 

$E_{distill}$ and $E_{search}$ are reasoning evidence mentioned in section \ref{sec:elabel}, while $E_{merge}$ is a combination of the two evidence labels in FeTaQA dataset and is the training labels for hightlighter. It's important to note that all three labels were obtained with knowledge of the golden summary $\mathcal{Y}$. 
According to Table \ref{tab:aba2}, the evaluation score corresponding to $E_{merge}$ is the highest, indicating that the evidence quality of $E_{merge}$ is the best.
This also indicates that although the overall quality of \( E_{search} \) obtained through greedy search is lower than \( E_{distill} \), the quality on some individual samples is higher than \( E_{distill} \).

\input{miscs/figure_cases}

\subsection{Cases Analysis}

We showcase some instances at of accurate and inaccurate evidence generated by HeLM's table highlighter, alongside outputs from HeLM-13B, TabCot(GPT3) and LLaMA2-13B-QLoRA respectively, as shown in Figure \ref{fig:cases}. For instance,  case (2) shows the results given two questions about numerical calculation. HeLM's highlighter accurately finds the player's records during their tenure at Coventry, aiding the table summarizer in precisely calculating the player's tenure and total appearances. In contrast, both LLaMA2-QLoRA and TabCot(GPT3) give wrong answers for the two questions.

Cases (3) and (4) represent instances where the highlighter made inaccurate judgments. In case (3), the highlighter highlighted two irrelevant rows, one of which appeared in the summarizer's output. In case (4), the highlighter missed highlighting one row, leading the summarizer to fabricate a ninth-ranking entry, but the table only contained data for the top eight ranks. 
Therefore, it's evident that the summarizer places significant emphasis on the highlighted segments of the table as identified by the highlighter. 

%% file: miscs/table_fetaqa.tex
\begin{table}[htbp]
  \centering
  \begin{tabular}{lp{7mm}p{7mm}p{7mm}p{8mm}} \toprule
        Models          & R-1  & R-2  & R-L  & BLEU    \\ \midrule
        \multicolumn{5}{c}{Fine-tuning based methods} \\ \midrule
        T5-small        & 55  & 33  & 47   & 21.60  \\
        T5-base         & 61  & 39  & 51   & 28.14  \\
        T5-large        & 63  & 41  & 53   & 30.54  \\
        UnifiedSKG      & 64  & 42  & 54   & 31.5  \\
        TAPEX           & 62  & 40  & 51   & 30.2  \\
        OmniTab         & 63  & 41  & 52   & 30.7  \\
        PLOG            & 64  & 43  & 55   & 31.8  \\
$\text{LLaMA2-13B}^{\dag}$&  \underline{66.5} &  44.7 &  \underline{56.2} & \underline{33.24}  \\
$\text{HeLM-13B}^{\dag}$& \textbf{67.8}  & \textbf{46.4}  & \textbf{57.9}   & \textbf{35.10}  \\ \midrule
        \multicolumn{5}{c}{Few-shot LLMs methods} \\ \midrule
        TabCot(GPT-3)   & 61  & 38  & 49   & 27.02  \\
        Dater(Codex)    & \underline{66}  & \underline{45}  & \underline{56}  & 30.92  \\ \bottomrule
  \end{tabular}
  \caption{Results on FeTaQA dataset. The $\dag$ marked models are trained using QLoRA.} 
  \vspace*{0.4cm}
  \label{tab:fetaqa}
\end{table}


%% file: miscs/table_qtsumm.tex
\begin{table}[h]
  \centering
  \begin{tabular}{lp{9mm}p{8mm}c} \toprule
        Models          & BLEU   & R-L    & METEOR    \\ \midrule
        \multicolumn{4}{c}{Fine-tuning based methods} \\ \midrule
        T5-Large        & 20.3   & 38.7   & 40.2  \\
        BART-large      & 21.2   & 40.6   & 43.0  \\
        OmniTab         & 22.4   & 42.4   & 44.7  \\
        TAPEX           & 23.1   & 42.1   & 45.6  \\
$\text{LLaMA2-13B}^{\dag}$&  \underline{23.3}  & \underline{42.8}   & 46.7 \\
$\text{HeLM-13B}^{\dag}$& \textbf{25.0}  & \textbf{45.3}   & \underline{50.0}  \\ \midrule
        \multicolumn{4}{c}{Few-shot LLMs methods} \\ \midrule
        LLaMA2-7B       & 14.0   & 31.2   & 37.3 \\
        LLaMA2-13B      & 17.5   & 33.2   & 42.3 \\
        LLaMA2-70B      & 19.0   & 38.0   & 46.4 \\
        GPT-3.5         & 20.0   & 39.9   & \underline{50.0}  \\
        GPT-4           & 19.5   & 40.5   & \textbf{51.1}  \\
 \bottomrule
  \end{tabular}
  \caption{Results on QTSumm dataset.} 
  \vspace*{0.4cm}
  \label{tab:qtsummm}
\end{table}


%% file: miscs/table_humanEval.tex
\begin{table}[htbp]\small
  \centering
  \begin{tabular}{lp{11mm}p{11mm}p{11mm}} \toprule
        Models          & Fluency & Correct & Adequate    \\ \midrule
        TabCot (GPT3)    & 2.05   & 1.98   & 2.02  \\
        LLaMA2-13B QLoRA    & 2.00   & 2.11   & 2.06  \\
        HeLM-13B           & \textbf{1.96}  & \textbf{1.92}   & \textbf{1.91} \\ \bottomrule
  \end{tabular}
  \caption{Human evaluation on FeTaQA. The numbers in the table indicate the average ranking.} 
  \vspace*{0.4cm}
  \label{tab:humaneval}
\end{table}


%% file: miscs/table_fetaqa_aba1.tex
\begin{table}[h]
  \centering
  \begin{tabular}{lp{9mm}p{9mm}p{9mm}p{12mm}} \toprule
        Models          & R-1  & R-2  & R-L  & BLEU    \\ \midrule
$\text{LLaMA2-7B}^{\dag}$&  65.0 &  43.0 &  54.8 & 32.68  \\
$\text{LLaMA2-13B}^{\dag}$&  66.5(\textcolor{red}{+0.5}) &  44.7(\textcolor{red}{+1.7}) &  56.2(\textcolor{red}{+1.4}) & 33.24(\textcolor{red}{+0.56})  \\ \midrule
$\text{HeLM-7B}^{\dag}$& 65.4 &  43.5 &  55.4 & 32.95  \\
$\text{HeLM-13B}^{\dag}$& 67.8(\textcolor{red}{+2.4}) &  46.4(\textcolor{red}{+2.9}) &  57.9(\textcolor{red}{+2.5}) & 35.10(\textcolor{red}{+2.15})  \\
         \bottomrule
  \end{tabular}
  \caption{Ablation study with different model size on FeTaQA dataset.} 
  \vspace*{0.4cm}
  \label{tab:aba1}
\end{table}

%% file: miscs/table_fetaqa_aba2.tex
\begin{table}[h]
  \centering
  \begin{tabular}{lp{11mm}p{11mm}p{11mm}p{12mm}} \toprule
        Models          & R-1  & R-2  & R-L  & BLEU   \\ \midrule
$\text{HeLM-13B}^{\dag}$& 67.8 &  46.4 &  57.9 & 35.10  \\
         - w/o highlight& 66.6($\text{\textcolor{gg}{-1.2}}$)   &  44.7($\text{\textcolor{gg}{-1.7}}$) &  56.6($\text{\textcolor{gg}{-1.3}}$) & 33.13($\text{\textcolor{gg}{-1.97}}$)\\
         - subTab       & 65.0($\text{\textcolor{gg}{-2.8}}$)   &  43.3($\text{\textcolor{gg}{-3.1}}$) &55.5($\text{\textcolor{gg}{-2.4}}$) & 32.28($\text{\textcolor{gg}{-2.82}}$)\\
         - $E_{distill}$  &  69.4(\textcolor{red}{+1.6}) & 47.8(\textcolor{red}{+1.4})&59.2(\textcolor{red}{+1.3}) &36.33(\textcolor{red}{+1.23})\\
         - $E_{search}$ & 68.1(\textcolor{red}{+0.3})    & 46.6(\textcolor{red}{+0.2})&58.0(\textcolor{red}{+0.1}) &34.96($\text{\textcolor{gg}{-0.14}})$\\
         - $E_{merge}$  &  69.6(\textcolor{red}{+1.8})   & 48.2(\textcolor{red}{+1.8})&59.5(\textcolor{red}{+1.6}) &36.74(\textcolor{red}{+1.64})\\  
         \bottomrule
  \end{tabular}
  \caption{Ablation study on FeTaQA dataset with different evidence.} 
  \vspace*{0.4cm}
  \label{tab:aba2}
\end{table}

%% file: miscs/figure_cases.tex
\begin{figure*}[t]
    \centering
    \includegraphics[width=1.0\textwidth]{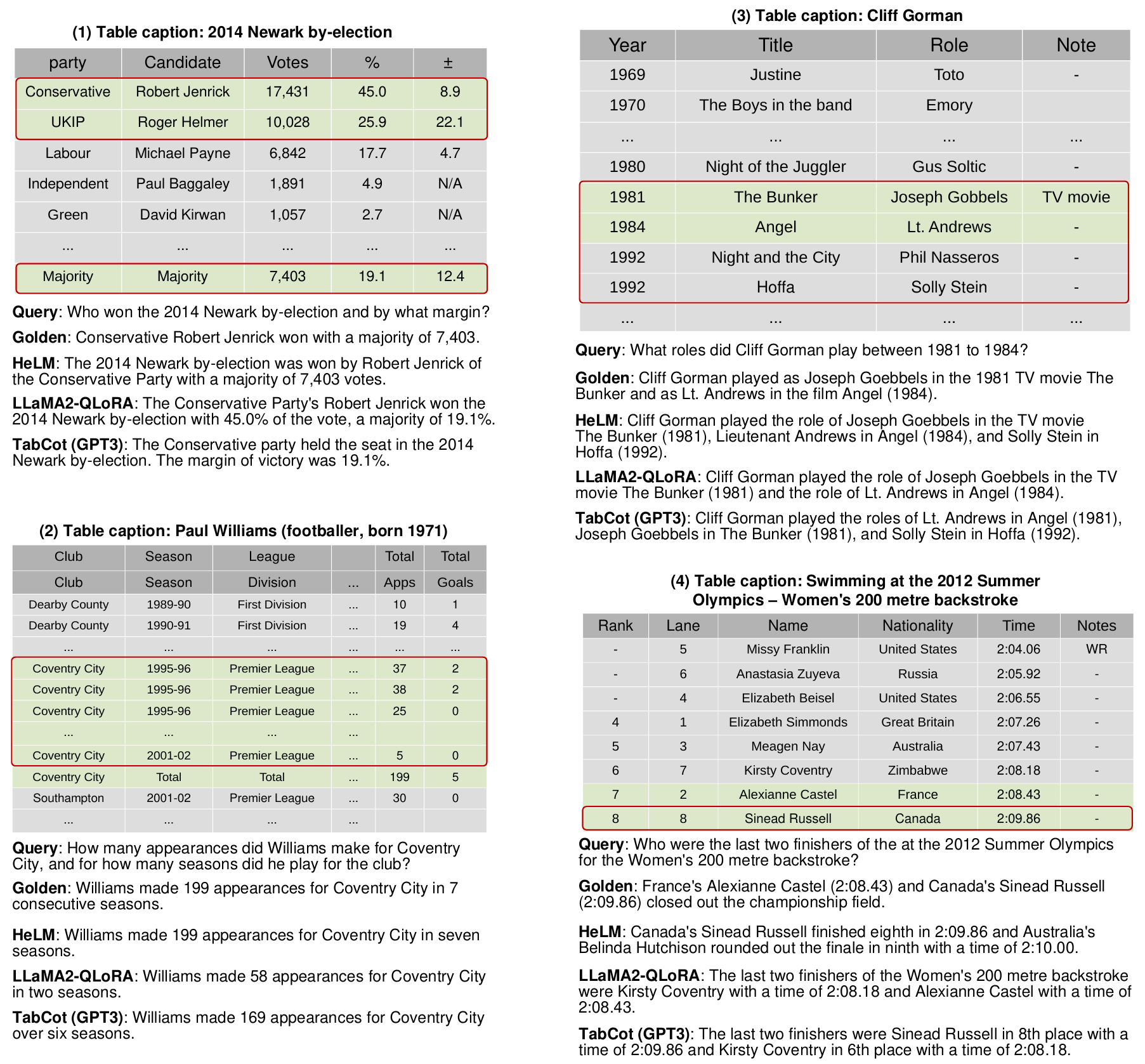}
    \caption{Cases from the FeTaQA Dataset. The highlighter of HeLM has highlighted specific parts of the table using red boxes. The rows in the table with a green background represent manually observed evidence related to the query.}
    \vspace*{0.6cm}
    \label{fig:cases}
\end{figure*}

%% file: sections/s5_conclusion.tex
\section{Conclusion}
\label{sec:conclude}
In this paper, leveraging existing open-source LLM, we devised a lightweight two-step table-to-text solution named HeLM. HeLM comprises two modules: table highlighter and table summarizer. Both modules adopt LLaMA2 as the backbone model and conduct efficient fine-tuning using designed prompts. Additionally, we explored diverse methods for constructing reasoning evidence, encompassing distillation from ChatGPT and construction by a searching algorithm. Our experimental findings showcase that leveraging the highlighter to highlight important row data of the input table significantly elevates the quality of the output and provides valuable interpretability.

 


Despite HeLM achieving good results on two table-to-text datasets, there are still some limitations and space for further improvement: (1) We haven't extensively investigated table highlighting formats, and there might be more effective ways. (2) Currently, HeLM is trained for specific datasets, lacking generalization; training HeLM on a mixture of table-to-text datasets could be a better solution. (3) The evidence labels generated by greedy search in table highlighter could be further improved. For instance, we can employ reinforcement learning to search for more optimal evidence labels. 
(4) Despite our model achieving high scores in BLEU and ROUGE metrics, its advantages in numerical and textual accuracy aren't notably pronounced compared to some powerful LLMs.




